\def\year{2022}\relax
\documentclass[letterpaper]{article} 
\usepackage{aaai22}  
\usepackage{times}  
\usepackage{helvet}  
\usepackage{courier}  
\usepackage[hyphens]{url}  
\usepackage{graphicx} 
\usepackage{natbib}  
\usepackage{caption} 
\DeclareCaptionStyle{ruled}{labelfont=normalfont,labelsep=colon,strut=off} 
\usepackage{algorithm}
\usepackage{algorithmic}
\usepackage{comment}

\usepackage{newfloat}
\usepackage{listings}
\usepackage{amsmath}
\floatstyle{ruled}
\newfloat{listing}{tb}{lst}{}
\floatname{listing}{Listing}

\usepackage{booktabs}
\usepackage{amsfonts}
\usepackage{enumitem}
\usepackage{color,soul}

\usepackage{hyperref}
\usepackage{xcolor}
\hypersetup{
    colorlinks = true,
    linkbordercolor = {white},
    linkcolor={cyan},
    urlcolor={blue},
    citecolor= black
}

\title{Reward Design For An Online Reinforcement Learning Algorithm Supporting Oral Self-Care}
\author {
    Anna L. Trella\textsuperscript{\rm 1},
    Kelly W. Zhang\textsuperscript{\rm 1},
    Inbal Nahum-Shani\textsuperscript{\rm 2},
    Vivek Shetty\textsuperscript{\rm 3},
    Finale Doshi-Velez\textsuperscript{\rm 1},
    Susan A. Murphy \textsuperscript{\rm 1}
}
\affiliations {
    \textsuperscript{\rm 1} Department of Computer Science, Harvard University\\
    \textsuperscript{\rm 2} Institute for Social Research, University of Michigan \\
    \textsuperscript{\rm 3}
    Schools of Dentistry \& Engineering, University of California, Los Angeles \\
    annatrella@g.harvard.edu, kellywzhang@seas.harvard.edu,  inbal@umich.edu, vshetty@ucla.edu, finale@seas.harvard.edu, samurphy@fas.harvard.edu
}

\usepackage{bibentry}

\begin{document}

\maketitle

\begin{abstract}
Dental disease is one of the most common chronic diseases despite being largely preventable. However, professional advice on optimal oral hygiene practices is often forgotten or abandoned by patients. Therefore patients may benefit from timely and personalized encouragement to engage in oral self-care behaviors. In this paper, we develop an online reinforcement learning (RL) algorithm for use in optimizing the delivery of mobile-based prompts to encourage oral hygiene behaviors. One of the main challenges in developing such an algorithm is ensuring that the algorithm considers the impact of the current action on the effectiveness of future actions (i.e., delayed effects), especially when the algorithm has been made simple in order to run stably and autonomously in a constrained, real-world setting (i.e., highly noisy, sparse data). We address this challenge by designing a quality reward which maximizes the desired health outcome (i.e., high-quality brushing) while minimizing user burden. We also highlight a procedure for optimizing the hyperparameters of the reward by building a simulation environment test bed and evaluating candidates using the test bed. The RL algorithm discussed in this paper will be deployed in Oralytics, an oral self-care app that provides behavioral strategies to boost patient engagement in oral hygiene practices.
\end{abstract}
\section{Introduction}
Although largely preventable, dental disease is one of the most common chronic diseases in the United States \cite{benjamin2010oral}. In addition to the personal pain and suffering and the financial costs of treating dental disease, oral health problems affect people's ability to eat and swallow, speak and socialize, and increase the risk of further health complications. Chronic oral infections have been associated with diabetes, heart and lung disease, stroke, and low birth weight. 
Dental disease is preventable through systematic, twice-a-day tooth brushing \cite{loe2000oral}. Yet, this basic behavior is not as widely practiced because patients forget or abandon clinician instructions \cite{weinstein1989effective,martin2005challenge}.

Ultimately, oral health depends on the individual's ability and willingness to carry out consistent brushing behaviors in the home setting. As innovation in healthcare works to move the field from focusing on costly reactive care for established diseases to proactive preventive care, there is increasing importance on promoting self-care. New approaches from behavioral science have the potential for modifying individual oral health behaviors \cite{buunk2011determinants}.
Rather than episodic clinician-delivered oral hygiene instruction, today's technologies can be leveraged to deliver engaging feedback and motivational messages to individuals in their home environments around the time they brush their teeth.

Towards that goal, our main objective is to build an online reinforcement learning (RL) algorithm \cite{sutton2018reinforcement} to be incorporated into the Oralytics mobile health application. Oralytics is a mobile health smartphone app that provides interventions to encourage oral self-care behaviors. The RL algorithm we develop will learn, online, users' responsivity to the messages and decide at each of two brushing times per day whether to deliver each user an intervention message; the goal of the algorithm is to improve users' brushing quality throughout the study. The first Oralytics clinical trial with an RL algorithm to optimize message delivery will begin in fall 2022.


The most commonly used RL algorithm in mobile health interventions are \textit{bandit algorithms} \cite{tewari2017ads,wang2005bandit,langford2007epoch}, one of the simplest types of RL algorithms. Bandit algorithms are commonly used because of their ability to run reliably and stably in an online environment, which is critical for high-stakes clinical trial settings. Mobile health clinical trials can require years of work by an interdisciplinary team to develop and the RL algorithm cannot be trivially changed once the study begins. So it is critical that the RL algorithm runs stably (e.g., the algorithm cannot take too long to update and fail to be ready to select actions each day) \cite{a15080255}.

However, classical bandit algorithms, designed to optimize the immediate reward, are not equipped to account for the delayed effects of actions (i.e., the impact of the current intervention on a user's responsivity to future interventions). Inappropriate (e.g., untimely or too many) notification interruptions may cause user burden, which can annoy users and cause them to be unresponsive. Preventing user burden is crucial, especially in mobile health settings, which typically have high user dropout rates \cite{meyerowitz2020rates, amagai2022challenges}. One option is to use RL algorithms that model a full Markov decision process (MDP), specifically by modeling how the action impacts the next state and the future rewards achievable when starting in that next state. However, because of the constrained setting of Oralytics, specifically, the highly-noisy outcomes and limited data per user, we do not expect to have enough data to 
effectively learn the large number of parameters needed in a MDP-based algorithm. Moreover, MDP-based algorithms may not be as stable to run and update online as bandit algorithms. Bandit algorithms can also be interpreted as a form of discount regularization for a full MDP-based RL algorithm \cite{jiang2015dependence}.

\textbf{Contributions:} In this work, we design a surrogate reward with a cost term that captures the negative, delayed effects of sending messages (Section~\ref{reward_design}). Note that we will continue to \textit{evaluate} our algorithms in terms of the original reward (true target), but will use surrogate rewards penalized by the cost term to optimize the \textit{learning} of the algorithm. Similar ideas have been used in the model predictive control literature \cite{jain2021optimal}. Additionally, we relate the cost term on the reward to a proxy for the impact of actions on the future state in the classical Bellman equation. Including such a cost term helps capture the delayed effects of actions in a simple way, allows us to continue to use bandit algorithms (which are stable and easier to run), and does not require the algorithm to learn additional parameters online. To evaluate the design of the cost term, we first build realistic simulation environment test beds using previously collected user brushing data (Section~\ref{available_data}). We then use both domain expertise and evaluation in these test beds to evaluate the design of the cost terms in their ability to capture the delayed negative effects of sending a message (Section~\ref{hyperparameter_opt}). An interesting takeaway from this work is that \textit{designing the rewards used by the algorithm is a critical component of real-world RL algorithm design.}

\section{Preliminaries}
The Oralytics RL algorithm will deliver feedback and motivational messages to users via notifications on a smartphone app. The first clinical trial of Oralytics with the RL algorithm will consist of approximately $N = 72$ users. Each user will participate in the study for 70 days. Each day there are two decision times, one hour before the user-specified morning and evening brushing times. This means for each user, there are a total of $T=140$ decision times. At each decision time $t \in [1, 140]$ for user $i$, the algorithm decides between action $A_{i, t} = 1$ (send a message) and $A_{i, t} = 0$ (do not send a message) given the user's current state $S_{i, t}$. After executing action $A_{i, t}$ in state $S_{i, t}$, a reward, $R_{i, t}$, is observed.   

The goal of this paper is to design the reward for the Oralytics algorithm, which optimizes for the desired health outcome while also preventing user burden, especially when needing to use a simple algorithm due to real-world constraints. The reward design will enhance the ability of the Oralytics algorithm to learn fast and select good actions stably in the real-world study. In the remainder of this section, we first discuss the available data from previous studies used to inform the design of the RL algorithm, and then we discuss the RL algorithm. In the following sections, we will discuss the design of the reward.

\subsection{Available Data from Previous Studies}
\label{available_data}

One of the main challenges is that we want to use available data to inform the design of the RL algorithm, however, the available data is sparse and only partially informative. We have access to two data sets on brushing behaviors, ROBAS 2 \cite{info:doi/10.2196/17347} and ROBAS 3. More importantly, both data sets only have observations under no action / intervention.

ROBAS 3 has a more sophisticated sensory suite (the same suite that will be used in Oralytics) than ROBAS 2. In addition to brushing duration, ROBAS 3 was able to collect sensory information such as brushing pressure to inform brushing quality, whereas ROBAS 2 was only able to record brushing duration. However, ROBAS 3 represents a slightly different population because the main goal of ROBAS 3 was to test the robustness of the new passive data collection system. More importantly, ROBAS 3 consisted of a fewer number of users with worse brushing performance (i.e., lower average brushing duration and more sessions with no brushing).
Table~\ref{data_set_differences} highlights additional key differences between the ROBAS 2 and ROBAS 3 data sets.

We use the ROBAS 2 data set to design the prior used in the RL algorithm. We use the ROBAS 3 data set to construct a quality simulation environment that serves as a test-bed for evaluating the RL algorithm candidates.

\begin{table}[H]
\begin{tabular}{p{0.50\linewidth}p{0.175\linewidth}p{0.175\linewidth}}
\toprule
Property & ROBAS 2 & ROBAS 3 \\ 
\midrule
Num. Users & 32 & 13 \\
Num. Datapoints & 1792 & 2188 \\
Intervention Messages & No & No \\
Sensor Data & No & Yes \\
Avg. Brushing Duration & 85.874 & 78.098 \\
\% Sessions with No Brushing & 38\% & 48\% \\
\bottomrule
\end{tabular}
\caption{Differences between the ROBAS 2 and ROBAS 3 data sets}
\label{data_set_differences}
\end{table}

\subsection{RL Algorithm}
The main goal of many RL problems is to maximize accumulated rewards. To accomplish this goal, online RL algorithms consist of (i) learning a model of the environment and (ii) an action selection strategy. In the case of Oralytics, we build off a contextual bandit framework which (i) learns a reward approximating function (model for the mean reward given the current state and action, $\mathbb{E}[R_{i, t} | S_{i, t}, A_{i, t}]$), and (ii) uses the learned model to select actions based on the current state $S_{i, t}$. 

Pursuing a full MDP-based RL algorithm involves modeling both state transition probabilities and the expected reward for each state and action to form a policy. As discussed in the introduction, due to the constrained setting of Oralytics, we do not expect to have enough data to effectively learn parameters for a full MDP-based algorithm. Thus, we modify a contextual bandit algorithm---specifically, we use a variant of linear posterior sampling \citep{DBLP:journals/corr/0001RKO17}. Posterior sampling involves using a Bayesian model for the mean reward and selection actions according to the posterior probability that each action is optimal. The Bayesian framework allows us to incorporate data from previous studies and domain knowledge into the prior distribution. Further, the actions are selected probabilistically, which facilitates after-study inference \cite{pmlr-v149-yao21a, zhang2020inference}.

For our posterior sampling algorithm, we use Bayesian Linear Regression (BLR) with action centering for our reward approximating function \cite{DBLP:journals/corr/abs-1909-03539}. BLR with action centering has a closed-form posterior update, only requires access to treatment effect features at decision times, and is robust to miss-specification of the correctness of the baseline reward model \cite{DBLP:journals/corr/abs-1909-03539}. 
Specifically, the BLR with action centering posterior sampling algorithm uses the following model of the reward:
\begin{multline}
    \label{eqn:blr}
    R_{i, t} = m(S_{i, t})^T \alpha_0 + \pi_{i,t} f(S_{i, t})^T \alpha_1 \\ 
    + (A_{i, t} - \pi_{i, t}) f(S_{i, t})^T \beta + \epsilon
\end{multline}
where $\pi_{i,t}$ is the probability that the RL algorithm selects action $A_{i,t} = 1$ for user $i$ in state $S_{i,t}$, and $\epsilon \sim \mathcal{N}(0, \sigma^2)$. We put a $\mathcal{N}(\mu_{\alpha_0}, \Sigma_{\alpha_0})$ prior on $\alpha_{0}$, a $\mathcal{N}(\mu_{\alpha_1}, \Sigma_{\alpha_1})$ prior on $\alpha_{1}$, and a $\mathcal{N}(\mu_{\beta}, \Sigma_{\beta})$ prior on $\beta$. We discuss the procedure for setting prior values for $\sigma^2, \mu_{\alpha_0}, \Sigma_{\alpha_0}, \mu_{\beta}, \Sigma_{\beta}$ using ROBAS 2 data in Appendix~\ref{fitting_prior}. 

Finally, note that we design the Oralytics RL algorithm to learn by pooling the data of all users in the study. Algorithms that learn using the data of all users have the potential to learn faster than those that learn using the data of a single user. Moreover, in early experiments, BLR with full pooling performed better than BLR that learns using only a single user's data or a smaller cluster of users' data \cite{a15080255}. Note that in Equation~\eqref{eqn:blr}, the reward model parameters are not indexed by the user $i$ to reflect how we are learning a single RL algorithm for all users in the study. For further details on the RL algorithm such as how the algorithm performs action selection and posterior updating, please see Appendix~\ref{app:rl_alg}.


\section{Reward Design for the Oralytics Algorithm}
\label{reward_design}

We evaluate our Oralytics RL algorithm in terms of its ability to maximize each user's total brushing quality $\sum_{t=1}^{T} Q_{i, t}$, where $Q_{i,t}$ is a non-negative measure of brushing quality observed after each decision time (two times a day).
Specifically, in collaboration with domain scientists on our team, we chose $Q_{i, t} = \min(B_{i, t} - P_{i, t}, 180)$, where $B_{i, t}$ is the user's brushing duration in seconds, and $P_{i, t}$ is the aggregated duration of over pressure in seconds. $Q_{i, t}$ is truncated to be at most $180$ to avoid optimizing for over-brushing.  $B_{i, t}$ and $P_{i, t}$ capture brushing quality as healthy brushing behaviors consist of brushing the dentist-recommended 120 seconds and using less pressure when brushing. For further discussion of the definition of brushing quality, please see Appendix~\ref{brush_quality_def}.

We now discuss the design of the reward that will be used by the RL algorithm. Throughout, we are interested in optimizing for brushing quality $Q_{i, t}$ and refer to $R_{i,t}$, the reward used by the RL algorithm, as the \textit{surrogate reward}.
The RL uses the surrogate rewards $R_{i, t}$ to update its parameters.
Specifically, let $R_{i, t} \in \mathbb{R}$ denote the surrogate reward for the $i$th user at decision time $t$:
\begin{equation}
\label{reward}
        R_{i, t} := Q_{i, t} - C_{i, t}
\end{equation}


Conceptually, the cost term is designed to allow the RL algorithm to optimize for healthy brushing behavior, while simultaneously considering the effect of the current message on the effectiveness of future interventions. Based on domain knowledge, we believe that sending a message at a decision time $t$ can only have a non-negative effect on the user's immediate brushing, $Q_{i,t}$. However, sending too many messages can risk habituation or may burden the user thus affecting user responsivity to future messages, i.e., affecting $Q_{i,t+1}, Q_{i,t+2}, \dots, Q_{i,T}$. 
Therefore, to anticipate these negative delayed effects of sending a message, we reduce the algorithm's reward when negative delayed effects are likely to occur.  $C_{i, t}$ provides this reduction as including $C_{i, t}$ in the algorithm's reward will provide a signal that sending a message ($A_{i, t} = 1$) may negatively affect future states. This signal is needed because we are using a contextual bandit type RL algorithm that does not explicitly model delayed effects of actions. 

In fact, $C_{i, t}$ can be viewed as a crude proxy for the delayed effect of actions in the Bellman equation in a MDP environment. Recall that according to the Bellman equation, it is optimal to select action $1$ over action $0$ if the immediate expected reward received from action $1$ over action $0$ exceeds the difference in optimal ``future values'' of selecting action $0$ over action $1$; specifically action $1$ and action $0$ can differ in ``future value'' due to the probability that each action will lead to a favorable or less favorable next state. Mathematically, this difference in future value is $\mathbb{E}[V^*(S_{i, t + 1}) | S_{i, t}, A_{i, t} = 0] - \mathbb{E}[V^*(S_{i, t + 1}) | S_{i, t}, A_{i, t} = 1]$ where $V^*$ is the optimal value function in a MDP setting. Note that in a pure contextual bandit setting the difference in future values of two actions is always zero, i.e., there are no delayed effects of actions because actions do not influence state transition probabilities. By including a cost on selecting action $1$, we can move from an always zero model of delayed effects of sending a message (selecting action $1$) used by the contextual bandit algorithm, to a more realistic setting in which there is some non-negative delayed effect of sending a message, captured by our cost term  $C_{i,t}$. For a full derivation, please see Appendix~\ref{relation_to_bellman}.

We define $\bar{B}_{i,t} := c_{\gamma}\sum_{j = 1}^{14} \gamma^{j-1} Q_{i, t - j}$ and $\bar{A}_{i,t} := c_{\gamma}\sum_{j = 1}^{14} \gamma^{j-1} A_{i, t - j}$. We set $\gamma = \frac{13}{14}$ to represent looking back 14 decision time points and scale each sum by constant $c_{\gamma}=\frac{1-\gamma}{1-\gamma^{14}}$ so that the weights sum to 1. Notice that our choice of $\gamma$ and the scaling constant means $0\le\bar{B}_{i,t}\le 180$ and $0\le \bar{A}_{i,t}\le 1$.
$\bar{B}_{i,t}$ captures the user's exponentially discounted brushing quality in the past week. $\bar{A}_{i,t}$ captures the number of interventions that were sent recently. 
Both terms are exponentially-discounted because we expect that interventions sent and user brushing in the near past will be more predictive of the delayed impact of the actions (via burden and/or habituation) than interventions sent and user brushing in the further past.

We define the cost of sending message (i.e. captures user burden in sending an intervention message) as:
\begin{equation}
\label{cost_term}
C_{i, t} := 
\begin{cases}
\xi_1 \mathbb{I}[\bar{B}_{i, t} > b] \mathbb{I}[\bar{A}_{i, t} > a_1] & \\
\hspace{10mm} + \xi_2 \mathbb{I}[\bar{A}_{i, t} > a_2]  & \smash{\raisebox{1.6ex}{if $A_{i, t} = 1$}} \\
0 & \hspace{-0mm} \mathrm{if~} A_{i, t} = 0
\end{cases}
\end{equation}

Notice that the algorithm only incurs a cost if the current action is to send an intervention, i.e., $A_{i, t} = 1$.  The first term $\xi_1 \mathbb{I}[\bar{B}_{i, t} > b] \mathbb{I}[\bar{A}_{i, t} > a_1]$ encapsulates the belief that if a high-performing user was sent too many messages within the past week, then we want to penalize the reward. The second term $\xi_2 \mathbb{I}[\bar{A}_{i, t} > a_2]$ encapsulates the belief that regardless of user performance, if they received too many messages within the past week, then we also want to penalize the reward. $b, a_1, a_2$ are chosen by domain experts. Notice that $a_1 < a_2$ because we believe a high-performing user will have a lower threshold of being annoyed by a message. The scientific team decided to set the following values:

\begin{itemize}
    \item $b=111$, is set to the 50th-percentile of user brushing durations in ROBAS 2, where brushing durations are truncated to 120 seconds if they exceed 120. 
    \item $a_1 = 0.5$, represents a rough approximation of the user getting a message 50\% of the time (rough approximation because we are using an exponential average mean) 
    \item $a_2 = 0.8$, represents a rough approximation of the user getting a message 80\% of the time (rough approximation because we are using an exponential average mean) 
\end{itemize}
$\xi_1, \xi_2$ are non-negative hyperparameters that we tune in Section~\ref{hyperparameter_opt}.
\section{Related Work}

\subsection{Reward Design }
As discussed in \citet{dulac2021challenges} reward design is one of the foremost challenges in real-world reinforcement learning. Specifying a quality reward is an important task because the reward design impacts the RL algorithm's ability to optimally select actions corresponding to the desired goal and the speed at which it learns \cite{mataric1994reward}. There are a variety of ways to modify and construct rewards. For example, one may design rewards to encourage RL algorithms to learn faster by encoding domain information, e.g., information about the relative effectiveness of different actions \cite{ng1999policy,laud2003influence,laud2004theory}. Additionally, in practice, rewards have been penalized to account for the cost of taking certain actions \cite{piette2022artificial} and to satisfy domain-specific constraints \cite{tessler2018reward}.

The idea of designing a surrogate reward that rather than using the true target as the reward that we utilize in this work is similar to the approach by \citet{jain2021optimal} to optimize a cost function using model predictive control (MPC). As discussed there, using MPC with the true cost function may not be the best choice due to errors in a dynamics model from the truncated planning horizon and the model approximations used. Their work shows that an MPC algorithm that uses a surrogate cost function (reward function) can have a lower true cost because it can mitigate problems that occur when the dynamics model is only approximate. Similarly in our setting,  $\sum_{t=1}^T Q_{i, t}$ is the true target, but using $Q_{i, t}$ as the reward may not be the best choice in cases where the RL algorithm must be made simple due to real-world constraints (e.g. managing high noise by using a contextual bandit algorithm rather than a full MDP based RL algorithm and using an approximate model for the reward function). Specifically, we penalize the reward with a cost term which,  when used by the online RL algorithm, should lead to a higher value of the true target, $\sum_{t=1}^T Q_{i, t}$.






\subsection{RL In Mobile Health}

RL algorithms are increasingly used in mobile health studies, for example, in studies to increase users' physical activity \citep{yom2017encouraging,figueroa2021adaptive,DBLP:journals/corr/abs-1909-03539,zhou2018personalizing,wang2021reinforcement}, to promote users' weight loss \citep{forman2019can}, and to help users' manage mental illness \citep{piette2022artificial}. Many of these works use contextual bandit algorithms, a popular choice for optimizing intervention delivery under real-world constraints discussed in \citet{a15080255} and \citet{figueroa2021adaptive}. Alternative algorithms include \citet{DBLP:journals/corr/abs-1909-03539} which uses a modified version of a posterior sampling contextual bandit algorithm, \citet{wang2021reinforcement} which uses an MDP framework and learns an action selection strategy offline, and \citet{zhou2018personalizing} which uses inverse RL to estimate a model of the reward and mixed-integer linear programming to select a 7-day schedule of personal step count goals for each user. Similar to this work, \citet{DBLP:journals/corr/abs-1909-03539} incorporated a term to approximate the delayed effects of the current action on future rewards. However, their approach did not involve explicitly developing a surrogate reward and they additionally did not allow the term modeling the delayed effects of actions to depend on previous user outcomes (only on previous actions). Finally, they also did not develop simulation environments with delayed effects to evaluate the design of their delayed effects of actions and do not provide explicit advice on how to choose the hyperparameters in their delayed effects term.





To the best of our knowledge, Oralytics is the first mobile health study utilizing an RL algorithm designed to prevent dental disease by supporting oral self-care behaviors by optimizing the delivery of motivational messages. Previous work \cite{a15080255} discusses a preliminary design of the RL algorithm for Oralytics. However, the main contribution of that paper was offering a generalizable framework for designing and evaluating an RL algorithm for digital interventions, by generalizing the predictability, computability, and stability (PCS) framework of \citet{yu2020veridical}. In this work, we follow the PCS framework developed in \citet{a15080255}. Although some of the challenges in Oralytics were tackled in the case study of that paper, reward design was not, which we address in this paper. Moreover, this paper takes advantage of new data, specifically the ROBAS 3 dataset, and we use the collected ROBAS 2 dataset to form the priors in our RL algorithm.



\section{Optimizing Hyperparameters in the Surrogate Reward}
\label{hyperparameter_opt}
Recall in Equation~\eqref{cost_term}, $\xi_1, \xi_2$ are hyperparameters that must be selected. We evaluate different values of $\xi_1, \xi_2$ in terms of their ability to maximize user brushing quality scores, $\sum_{t=1}^T Q_{i, t}$, across a variety of plausible environments. To do this, in each simulation environment variant, we perform a grid search over the range of possible values of $\xi_1, \xi_2$. Specifically for each algorithm variant (different values of $\xi_1, \xi_2$) and simulation environment pair, we consider evaluating two criteria:

\begin{enumerate}
    \item Average cumulative brushing quality across all users, $\frac{1}{N} \sum_{i=1}^N \sum_{t=1}^T Q_{i, t}$
    \item 25th-percentile of cumulative brushing quality, $\sum_{t=1}^T Q_{i, t}$, across all users
\end{enumerate}
We use these metrics to evaluate algorithm performance on both average and worse-off users.

\subsection{Simulation Environment Design}
Our simulation environment design involves three main components: (i) a base model of the environment under no intervention, (ii) initial treatment effect sizes for each user, and (iii) a procedure for modeling delayed effects of actions, specifically by shrinking users' treatment effect sizes.

\paragraph{Base Model of Environment}
Recall that we use the ROBAS 3 data set to construct the simulation environment. Even though ROBAS 3 did not involve intervention messages, we can still use the dataset to inform the base model for the simulation environment (i.e., a model for generating brushing quality under no action). For the base model, we fit a single zero-inflated Poisson model \cite{feng2021comparison} per user in the ROBAS 3 study. Each zero-inflated model's Bernoulli component represents a latent state of the user's intention to brush and the Poisson component represents brushing quality when the user intends to brush. We use a zero-inflated model because brushing quality is a highly zero-inflated outcome; many users miss brushing sessions altogether and have brushing quality zero \cite{a15080255}. 

\paragraph{Initial User Treatment Effects}
A user's responsiveness to an intervention is measured by their unique treatment effect sizes $\Delta_{i, B}$ and $\Delta_{i, N}$. $\Delta_{i, B}$ is the imputed treatment effect for the Bernoulli component and $\Delta_{i, N}$ is the imputed treatment effect for the Poisson component. The larger the effect sizes, the greater the user's responsiveness to the intervention. Since ROBAS 3 had no data under intervention, we impute $\Delta_{i, B}$ and $\Delta_{i, N}$ for each user by drawing a value from a zero-truncated normal distribution with mean and variance informed by the fitted parameters of the environment base model for that user $w_{i,b}$ and $w_{i,p}$. Further details on building the base simulation environment and imputing effect sizes can be found in Appendix~\ref{app:sim_env}. 

Incorporating the base simulation model and the effect sizes, the environment generates brushing quality $Q_{i, t}$ under action $A_{i, t}$ in state $S_{i, t}$ using:

$$
Z_{i,t} \sim \text{Bern} \left(1 -  \widetilde{p}_{i, t}\right)
$$
$$
\widetilde{p}_{i, t}=\mathrm{sigmoid} \left(g(S_{i, t})^Tw_{i,b} - A_{i, t} \max(h(S_{i, t})^T\Delta_{i,B}, 0)\right)
$$
$$
Y_{i,t} \sim \text{Pois} \left(\lambda_{i, t} \right)
$$
$$
\lambda_{i, t} = \exp \left( g(S_{i, t})^Tw_{i,p} + A_{i, t}\max(h(S_{i, t})^T\Delta_{i,N}, 0) \right)
$$
$$
Q_{i, t} = Z_{i,t} Y_{i,t}
$$
Above, $g(S_{i, t})$ is the baseline feature vector and $h(S_{i, t})$ is the feature vector that interacts with the effect size found in Appendix~\ref{baseline_features} and Appendix~\ref{treatment_features}.

\paragraph{Modeling Delayed Effects of Actions}
We model delayed effects of actions by shrinking users' responsiveness to interventions, i.e., their initial treatment effect sizes. Specifically, we shrink users' effect sizes by a factor $E \in (0,1)$ when a certain ``habituation'' criterion is met. Specifically the habituation criterion if either of the two scenarios hold: (a) $\mathbb{I}[\bar{B}_{i, t} > b]$ (user brushes well) and $\mathbb{I}[\bar{A}_{i, t} > a_1]$ (user was sent too many messages for a healthy brusher), or (b) $\mathbb{I}[\bar{A}_{i, t} > a_2]$ (the user has been sent too many messages). The first time a users' habituation criterion has been met, the user's future effect sizes $\Delta_{i, B}, \Delta_{i, N}$ starting at time $t + 1$ will be scaled down proportionally by $E$ for some $E \in (0,1)$ (the user is less responsive to treatment). Then after a week, at time $t + 14$, we will check the habituation criterion again. If the habituation criterion is met again, the effect sizes will be further shrunken by a factor of $E$ down to $E^2\cdot\Delta_{i,B}, E^2\cdot\Delta_{i,N}$ starting at time $t + 15$. However, if the habituation criterion is not fulfilled, then the user recovers their original effect size $\Delta_{i,B}, \Delta_{i,N}$ starting at time $t + 15$. 
This procedure continues until the user finishes the study. Notice that this means the user can only have their effect size shrunk at most once a week. This procedure simulates how the user may experience habituation (reduction in perception of the messages), but after a week, if the RL algorithm does not intervene too much, the user may dis-habituate and recover their prior responsivity. 

\subsection{Experiments and Results}

\begin{figure}[h]
    \centering
    {\includegraphics[width=0.23\textwidth,clip]{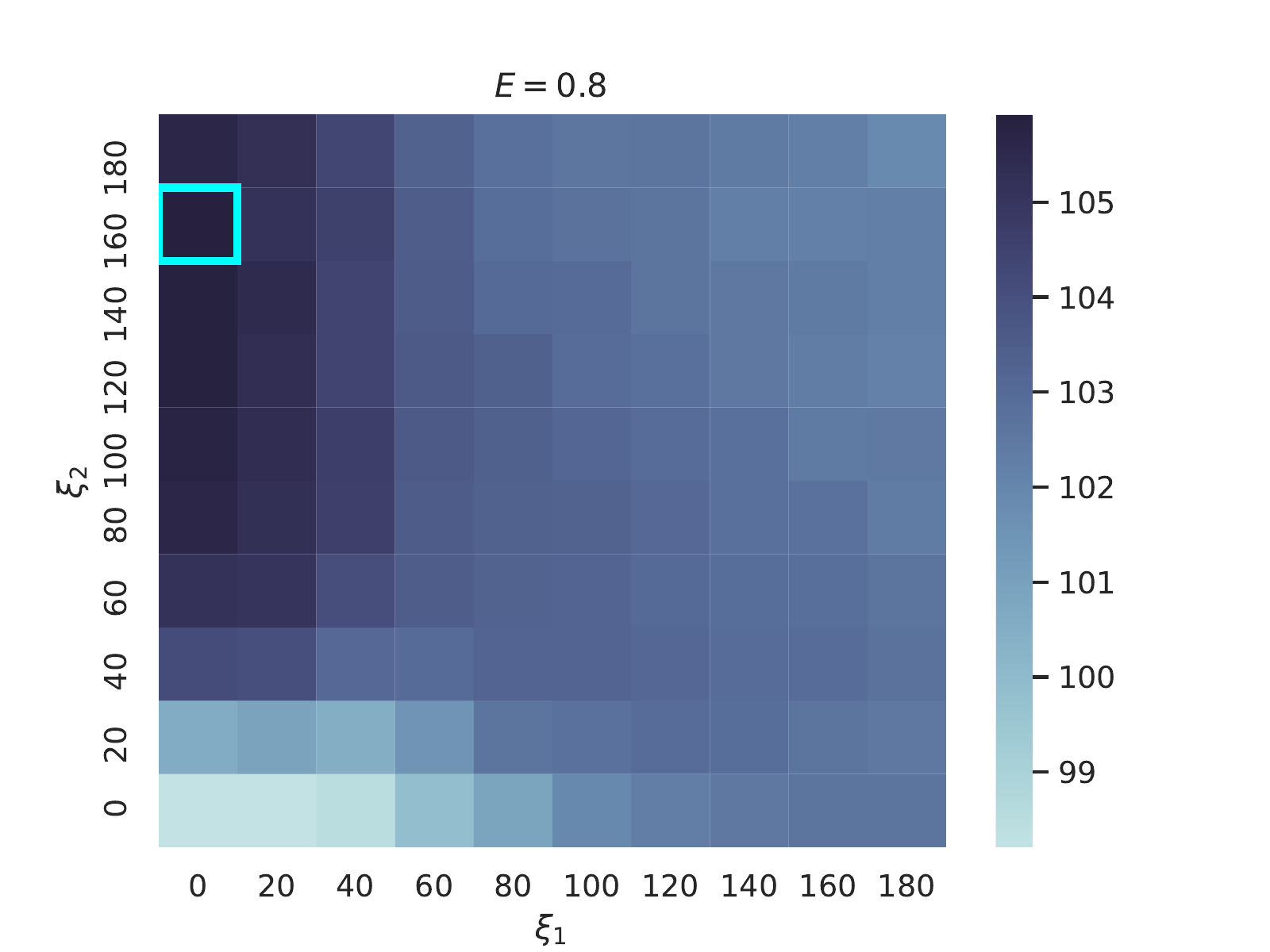}}
    {\includegraphics[width=0.23\textwidth,clip]{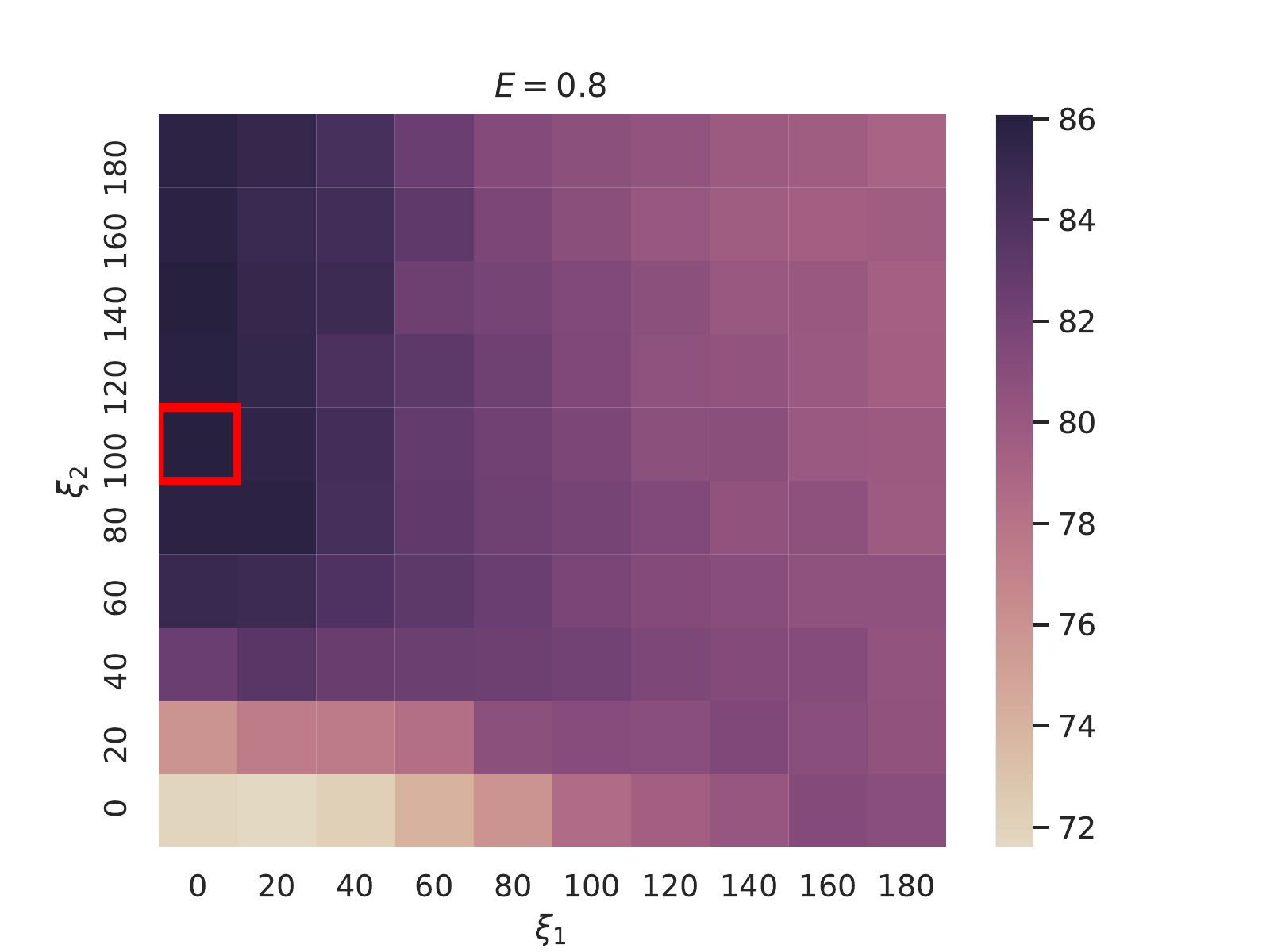}}
    {\includegraphics[width=0.23\textwidth,clip]{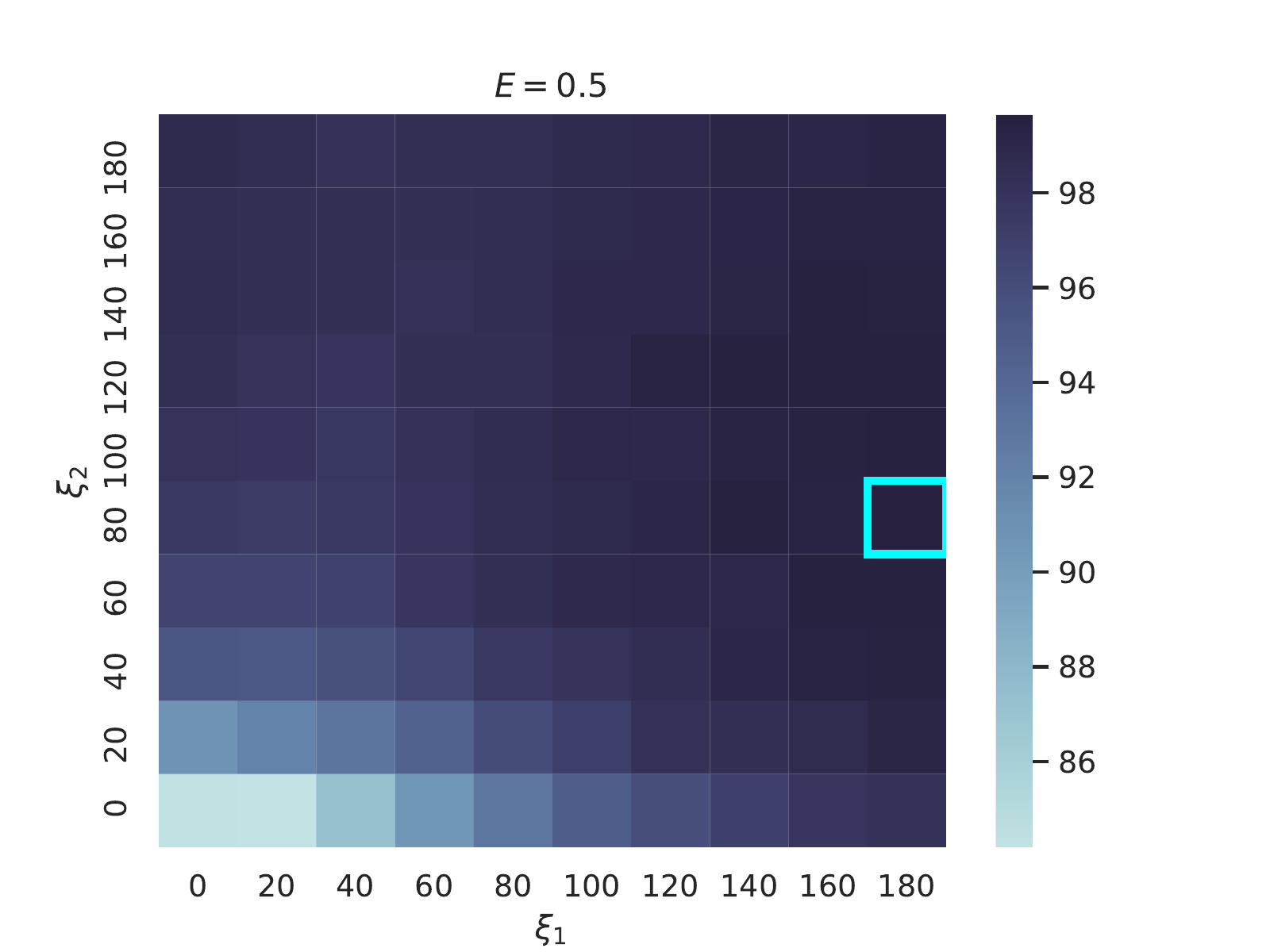}}
    {\includegraphics[width=0.23\textwidth,clip]{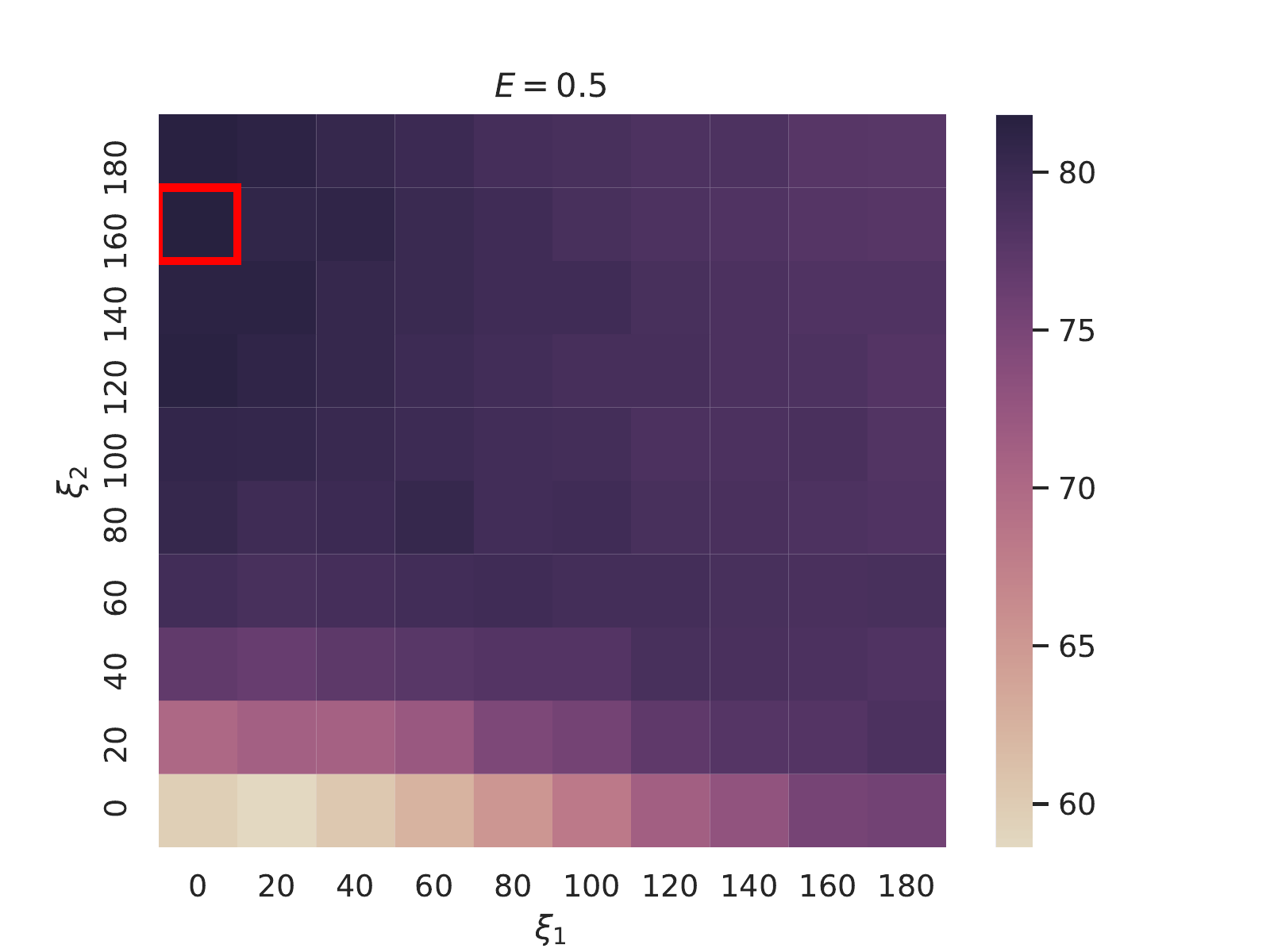}}
    {\includegraphics[width=0.23\textwidth,clip]{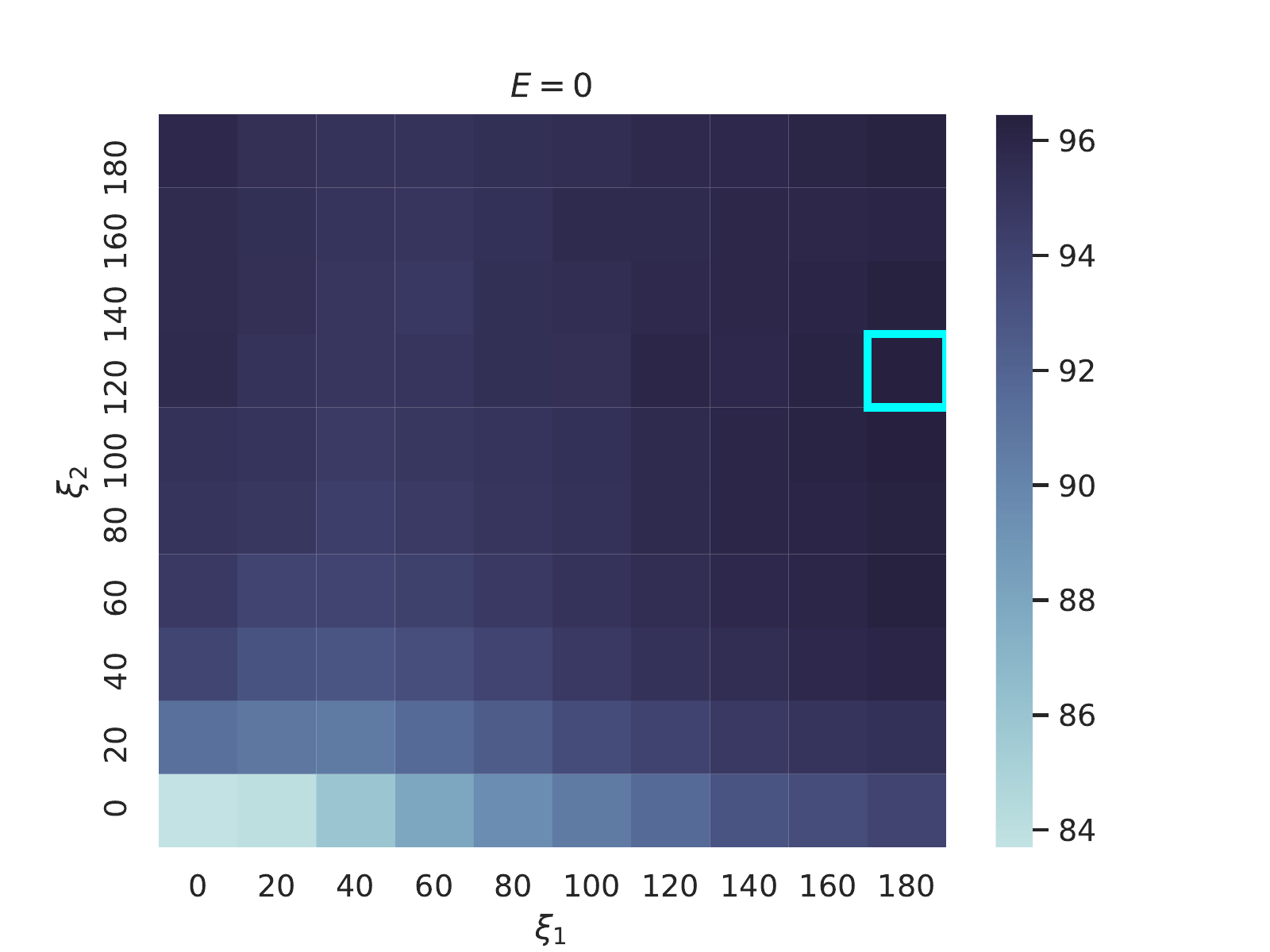}}
    {\includegraphics[width=0.23\textwidth,clip]{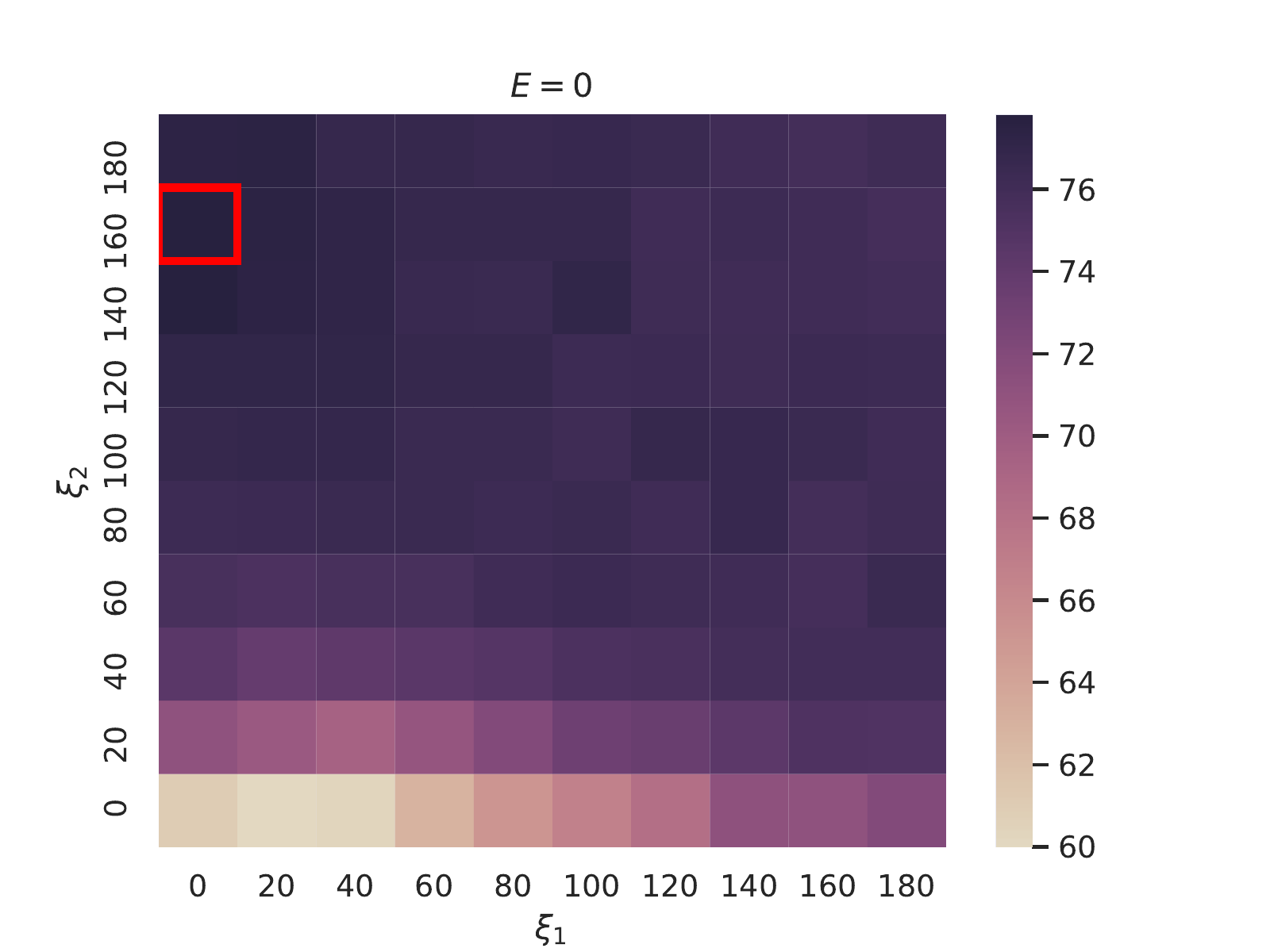}}
    \caption{\textbf{Heatmap of Candidate Values for $\xi_1, \xi_2$.} 
    We evaluate candidate values $\xi_1, \xi_2$ using two metrics and across three simulation environment variants. 
    Effect size shrinkage $E$ varies from slight shrinkage ($E=0.8$) to severe shrinkage ($E=0$) from top to bottom, The left column (blue) shows grids evaluated using average $\sum_{t=1}^{T} Q_{i, t}$ across users and the right column (purple) shows grids evaluated using $25$th-percentile of $\sum_{t=1}^{T} Q_{i, t}$ across users. The grid with the highest criteria value is boxed for readability.
    }
    \label{figs/heatmaps}
\end{figure}

We consider 3 values for shrinking the effect size, $E \in \{0, 0.5, 0.8\}$, for a total of 3 environment variants.
For each environment and RL algorithm (determined by the $\xi_1, \xi_2$ candidate value pairing), we simulate a study with $N=72$ users over $T=140$ decision times. The $N$ users are drawn with replacement from all the user environment models fitted using ROBAS 3 data. Since the Oralytics study is expected to recruit users into the study at a rate of about four users per week, we also simulate incremental recruitment by having 4 users enter the study per week. Following the procedure described above, for each environment variant, we generate two grids corresponding to the two evaluation criteria described at the beginning of this section (Section \ref{hyperparameter_opt}). Each square in a grid represents a criterion evaluated on a simulated study using values $\xi_1, \xi_2$ for the cost term, averaged across $100$ Monte Carlo simulated trials. Figure \ref{figs/heatmaps} shows heat maps of evaluation criterion values for different values of $\xi_1, \xi_2$ for each of the three values of effect size shrinkage, $E$, we consider.

The primary takeaway from our simulation results from Figure \ref{figs/heatmaps} is that using a surrogate reward is beneficial. This is because, on all of the heatmaps in Figure \ref{figs/heatmaps} below, the worst performing RL algorithm is that which learned with rewards with parameters $\xi_1 = 0$ and $\xi_2 = 0$, corresponding to learning using the original brushing qualities with a zero cost term. Thus, even though there are different optimal values of $\xi_1$ and $\xi_2$ for each environment and each evaluation criterion, it appears that choosing \textit{some} non-zero values of $\xi_1$ and $\xi_2$ is strictly beneficial.

Now consider the first evaluation criterion, the average cumulative brushing quality across all users, $\frac{1}{N} \sum_{i=1}^N \sum_{t=1}^T Q_{i, t}$. The left column of Figure~\ref{figs/heatmaps} shows results with different levels of delayed effects of actions. Notice that as the strength of the delayed effects increases, i.e., $E$ decreases, the penalty for burdening the users becomes harsher, and thus higher values of $\xi_1$ and $\xi_2$ are preferable. Note that for the $E=0.8$ case, we believe $\xi_1 = 0$ is favorable because in this environment the delayed effects of actions are relatively weak and the penalty incurred by the delayed effects is not outweighed by the benefit of sending users additional messages (recall that $\xi_1$ weights the condition where the user brushes well and was sent a moderate amount of messages).

Regarding evaluation criterion 2, the $25^{\mathrm{th}}$ percentile of $\frac{1}{T} \sum_{t=1}^T Q_{i,t}$ across users, notice that the optimal value of cost term hyperparameter $\xi_2$ increases with the severity of the delayed effects (lower values of $E$ means more severe delayed effects). The reason that $\xi_1 = 0$ is favored for this second evaluation criterion is that for these worse-off users, it was very rare for them to satisfy the condition that leads them to incur the cost due to the $\xi_1$ term.

To select $\xi_1, \xi_2$, we want to balance providing high value for the average and the worst-performing user. Other factors that influence our decision include prioritizing environment variants that the scientific team considers most important and considering the variation of grid values in the heat maps. For example, we may be willing to forgo an optimal value in a grid with low variation (values perform similarly) than an optimal value in a grid with high variation. Thus, based on all our simulation results, we plan to choose intermediate values of both $\xi_1$ and $\xi_2$, specifically $\xi_1 = 100$ and $\xi_2 = 100$. The scientific team will reconsider these values as we collect more ROBAS 3 brushing data.

\section{Discussion}
In this paper, we describe the reward design for developing an online RL algorithm that will be deployed in Oralytics, a mobile health app promoting oral self-care behaviors. By designing a surrogate reward to include a cost term, instead of having the RL algorithm learn using the true target, we generalize the contextual bandit framework to deal with the key challenge of capturing negative, delayed effects of interventions. After evaluating these reward candidate values in our simulation test-bed, we chose hyperparameter values of the reward that balances performance of the RL algorithm for the average user and the worst-off user across all environment variants. As our development for Oralytics is on-going, we will revisit the surrogate reward construction and will be in discussion with the scientific team in finalizing the algorithm that goes into the actual clinical trial.

\section{Acknowledgments}
This research was funded by NIH grants IUG3DE028723, P50DA054039, P41EB028242, U01CA229437, UH3DE028723, and R01MH123804. KWZ is also supported by the National Science Foundation grant number NSF CBET–2112085 and by the National Science Foundation Graduate Research Fellowship Program under Grant No. DGE1745303. Any opinions, findings, and conclusions or recommendations expressed in this material are those of the author(s) and do not necessarily reflect the views of the National Science Foundation.

\bibliography{main.bib}

\onecolumn
\appendix
\textbf{{\huge Appendices}}

\section{Data and Code}
We use two data sets from previous studies: ROBAS 2 \cite{info:doi/10.2196/17347} and ROBAS 3. The data sets are publically available \href{https://github.com/ROBAS-UCLA/ROBAS.2/blob/master/inst/extdata/robas_2_data.csv}{here} for ROBAS 2 and \href{https://github.com/ROBAS-UCLA/ROBAS.3/blob/main/data/robas_3_data.csv}{here} for ROBAS 3. Code for this paper can be found in a github repository  \href{https://github.com/StatisticalReinforcementLearningLab/oralytics_reward_design}{here}.

\section{Simulation Environments}
\label{app:sim_env}
\subsection{Defining the Target: Brushing Quality $Q_{i, t}$}
\label{brush_quality_def}
Brushing quality is the true target that the scientific team cares about in achieving healthy brushing behaviors (desired behavioral health outcome). $Q_{i, t}$ can be interpreted as a non-penalized reward and the true target capturing the health outcome that the scientific team's wants to maximize. The model of brushing quality we use is $Q_{i, t} = \min(B_{i, t} - P_{i, t}, 180)$. $B_{i, t}$ is the participant's brushing duration in seconds and $P_{i, t}$ is the aggregated duration of over pressure in seconds. $Q_{i, t}$ is truncated to avoid optimizing for over-brushing.

We also considered including other sensor information in our model of the brushing quality, including zoned brushing duration, a measure of how evenly users brush across the four zones (e.g. top-left quadrant, bottom-right quadrant, etc.). However, we did not end up including it in our brushing quality score because zoned brushing duration is not reliably obtainable as it requires Bluetooth connectivity and the participant to stand close enough to the docking station. Zoned brushing duration only appeared in about 82\% of brushing sessions in the pilot study.



\subsection{Building the Environment Base Model}
\label{env_base_model}

We build our base simulation environment using the ROBAS 3 dataset. We only use the ROBAS 2 dataset for fitting the prior of the algorithm. Our procedure for building the environment base model is similar to the environment discussed in \cite{a15080255}, however with the more advanced sensory suite in ROBAS 3, we consider brushing quality instead of just brushing duration. In addition, we only use a zero-inflated Poisson model to generate brushing quality and a non-stationary feature space.

\subsubsection{Baseline Feature Space of the Environment Base Models}
\label{baseline_features}
We form the following features using domain expert knowledge from behavioral health and dentistry. In this section we discuss how we fit a model for the baseline reward (i.e., the brushing quality under action $A_{i,t} = 0$). 
A discussion on how we model brushing quality under action $A_{i,t} = 1$ is in Appendix~\ref{imputing_eff_sizes}.

$g(S_{i,t}) \in \mathbb{R}^6$ denotes the feature space used to fit a model of the baseline reward which is the following:

\begin{enumerate}
    \item Bias / Intercept Term $\in \mathbb{R}$
    \item Time of Day (Morning/Evening) $\in \{0, 1\}$
    \item Prior Day Total Brushing Quality (Normalized) $\in \mathbb{R}$ 
    \item Weekend Indicator (Weekday/Weekend) $\in \{0, 1\}$
    \item Proportion of Non-zero Brushing Sessions Over Past 7 Days $\in [0, 1]$
    \item Day in Study (Normalized) $\in [-1,1]$
\end{enumerate}

\subsubsection{Normalization of State Features}
\label{normalizations}
We normalize features to ensure that all state features are within a similar range. The Prior Day Total Brushing Quality feature is normalized using z-score normalization (subtract mean and divide by standard deviation). Since each participant in the ROBAS 3 study had varying participant study lengths (due to dropping out), the Day in Study feature (originally in the range $[1:T_{i}]$ where $T_{i}$ represents the number of days user $i$ was in the study) is normalized to be between $[-1, 1]$. 
Note that when generating rewards, Day in Study is normalized based on Oralytic's anticipated 70 day study duration (range is still $[-1,1]$).
$$\text{Normalized Total Brushing Quality in Seconds} = (\text{Brushing Quality} - 154) / 163 $$

$$\text{Normalized Day in Study When Fitting Model for User } i 
$$
$$=\bigg (\text{Day} - \frac{T_{i} + 1}{2}\bigg ) / \frac{T_{i} - 1}{2}$$

$$\text{Normalized Day in Study When Generating Rewards} = (\text{Day} - 35.5) / 34.5$$

\subsubsection{Zero-Inflated Poisson Model for Brushing Quality}
\label{zero_infl_poisson}
Due to the zero-inflated nature of the ROBAS 3 data set, we use a zero-inflated Poisson model to generate brushing quality. $w_{i,b}, w_{i,p}$ are user-specific weight vectors, $g(S_{i, t})$ is the baseline feature vector of the current state defined in Appendix~\ref{baseline_features}, and $\mathrm{sigmoid}(x) = \frac{1}{1 + e^{-x}}$ is the sigmoid function.

To generate brushing quality for user $i$ at decision time $t$, we use the following model:
$$
Z \sim \text{Bernoulli}\left(1 - \mathrm{sigmoid}(g(S_{i, t})^T w_{i,b}) \right)
$$
$$
Y \sim \text{Poisson} \left( \exp \left( g(S_{i, t})^T w_{i,p} \right) \right)
$$
$$
\text{Brushing Quality} : Q_{i, t} = ZY
$$

\subsubsection{Fitting the Environment Base Model}
\label{fitting_env_base_models}
We use ROBAS 3 data to fit the brushing quality model under no intervention ($A_{i, t} = 0$). We fit one model per user and each user model was fit using MAP with a prior $w_{i,b}, w_{i,p} \sim \mathcal{N}(0, I)$ as a form of regularization because we have sparse data for each user. We jointly fit parameters for both the Bernoulli and the Poisson components. Weights were chosen by running random restarts and selecting the weights with the highest log posterior density.



\subsection{Imputing Treatment Effect Sizes for Simulation Environments}
\label{imputing_eff_sizes}
Recall that the ROBAS 3 data set does not have data under interventions (sending a message). Therefore we impute user-specific treatment effect sizes to model the reward under action 1. We use two guidelines to design the effect sizes following \cite{a15080255}:
\begin{enumerate}
    \label{effect_size_rationale}
    \item \label{effect_size_rationale:magnitude} For mobile health digital interventions, we expect the treatment effect (magnitude of weight) of actions to be smaller than (or on the order of) the effect of baseline features (baseline features are specified in Appendix~\ref{baseline_features}).
    \item \label{effect_size_rationale:variance} The variance in treatment effects across users should be on the order of the variance in the effect of features across users (i.e., variance in parameters of fitted user-specific models).
\end{enumerate}

We first construct a population-level effect size and then use the population effect size to sample unique effect sizes for each user. Following guideline \ref{effect_size_rationale:magnitude} above, to set the population level effect size, we first take the absolute value of the weights (excluding that for the intercept term) of the user base models fitted using ROBAS 3 data and then average across users and features (e.g., the average absolute value of weight for time of day). To generate user-specific effect sizes, for each user, we draw a value from a truncated normal centered at the population effect sizes. Following guideline \ref{effect_size_rationale:variance}, the variance of the truncated normal distributions is found by again taking the absolute value of the weights of the base models fitted for each user, averaging the weights across features, and taking the empirical variance across users. A more detailed procedure for imputing effect sizes is found in Appendix~\ref{procedure_eff_size_impute}.

\subsubsection{Treatment Effect Feature Space}
\label{treatment_features}
The treatment effect (advantage) feature space was made after discussion with domain experts on which features are most likely to interact with the intervention (action).
$h(S_{i, t}) \in \mathbb{R}^5$ denotes the features space used for the treatment effect which is the following:

\begin{enumerate}
    \item Bias/Intercept Term $\in \mathbb{R}$
    \item Time of Day (Morning/Evening) $\in \{0, 1\}$
    \item Prior Day Total Brushing Quality (Normalized) $\in \mathbb{R}$ 
    \item Weekend Indicator (Weekday/Weekend) $\in \{0, 1\}$
    \item Day in Study (Normalized) $\in [-1,1]$
\end{enumerate}

\subsubsection{Environment Model Including Effect Sizes}
We impute treatment effects on both the Bernoulli component which represents the user's intent to brush and the Poisson component which represents the user's brushing quality when they intend to brush. After incorporating treatment effects, brushing quality $Q_{i, t}$ under action $A_{i, t}$ in state $S_{i, t}$ is:

$$
Z \sim \text{Bernoulli} \bigg(1 - \mathrm{sigmoid} \big( g(S_{i, t})^\top w_{i,b} - A_{i, t} \cdot \max \big[ \Delta_{i,B} h(S_{i, t})^\top \boldsymbol{1}, 0 \big] \big) \bigg)
$$
$$
Y \sim \text{Poisson} \bigg( \exp \big( g(S_{i, t})^\top w_{i,p} + A_{i, t} \cdot \max\big[ \Delta_{i,N} h(S_{i, t})^\top \boldsymbol{1}, 0 \big] \big) \bigg)
$$
$$
Q_{i, t} = ZY
$$

Above we use $\boldsymbol{1}$ to refer to a vector of $1$'s, i.e., $\boldsymbol{1} = [1, 1, 1, 1, 1] \in \mathbb{R}^5$.
$\Delta_{i,B}, \Delta_{i,N}$ are user-specific effect sizes. $g(S_{i, t})$ is the baseline feature vector as described in Appendix~\ref{baseline_features}, and $h(S_{i, t})$ is the feature vector that interacts with the effect size specified above.

Notice that our design means the effect size on the Bernoulli component must be negative and the effect size on the Poisson component must be positive. If this is not the case, then that means in the current context, not sending a message will yield a higher brushing quality than sending a message. This is unreasonable because the only negative consequences of sending a message is the diminishing the responsivity to future messages. 
We ensure that $\max \big[ \Delta_{i,B} h(S_{i, t})^\top \boldsymbol{1}, 0 \big]$ and $\max\big[ \Delta_{i,N} h(S_{i, t})^\top \boldsymbol{1}, 0 \big]$ are non-negative to prevent the effect size from switching signs and having a negative effect on brushing quality.

\subsubsection{Procedure For Imputing Effect Sizes}
\label{procedure_eff_size_impute}
We consider a unique realistic effect size for each user. We first construct a population level effect size for the Bernoulli and Poisson components, $\Delta_B, \Delta_N$ respectively. We then use $\Delta_B, \Delta_N$ to sample user-specific effect sizes $\Delta_{i,B}, \Delta_{i,N}$.

Recall that for the environment base model, we fit a user-specific model for the brushing quality and obtained user-specific parameters $w_{i,b}, w_{i,p} \in \mathbb{R}^6$ (values of the fitted parameters can be found \href{https://github.com/StatisticalReinforcementLearningLab/oralytics_reward_design/tree/main/sim_env_data}{here}). Therefore we use the fitted parameters to form the population effect sizes as follows: 

\begin{itemize}
    \item $\Delta_B = \mu_{B,\text{avg}}$ where $\mu_{B,\text{avg}} = \frac{1}{5} \sum_{d \in [2 \colon 6]} \frac{1}{N} \sum_{i=1}^N |w_{i,b}^{(d)}|$.
    \item $\Delta_N = \mu_{N,\text{avg}}$ where $\mu_{N,\text{avg}} = \frac{1}{5} \sum_{d \in [2 \colon 6]} \frac{1}{N} \sum_{i=1}^N |w_{i,p}^{(d)}|$.
\end{itemize}

We use $w_{i,b}^{(d)}, w_{i,p}^{(d)}$ to denote the $d^{\mathrm{th}}$ dimension of the vector $w_{i,b}, w_{i, p}$ respectively; we take the average over all dimensions excluding $d=1$, which represents the weight for the bias/intercept~term.

Now to construct the user-specific effect sizes, we draw effect sizes for each each user from truncated normal distributions with support $[0, \infty)$:
$$
\Delta_{i,B} \sim \text{Truncated-Normal}_{[0, \infty)}(\Delta_{B}, \sigma^2_{B})
$$
$$
\Delta_{i,N} \sim \text{Truncated-Normal}_{[0, \infty)}(\Delta_{N}, \sigma^2_{N})
$$
We constrain the effect sizes to be non-negative to reflect how the only negative consequences of sending a message is the diminishing the responsivity to future messages. 

$\sigma^2_{B}, \sigma^2_{N}$ are the empirical standard deviation (SD) over the average of the fitted parameters and are set by the following procedure: 
\begin{itemize}
    \item $\sigma_B$ is the empirical SD over $\{ \mu_{i,B} \}_{i=1}^{N}$ where $\mu_{i,B} = \frac{1}{5} \sum_{d \in [2 \colon 6]} |w_{i,b}^{(d)}|$. 
    \item $\sigma_N$ is the empirical SD over $\{ \mu_{i,N} \}_{i=1}^{N}$ where $\mu_{i,N} = \frac{1}{5} \sum_{d \in [2 \colon 6]} |w_{i,p}^{(d)}|$.
\end{itemize}

After following the procedure described above, we found $\Delta_B = 0.743, \Delta_N=0.227$, $\sigma_B~=0.177$, and $\sigma_N = 0.109$ (values are rounded to the nearest 3 decimal places).

\section{RL Algorithm}
\label{app:rl_alg}
For the RL algorithm, we use a contextual bandit algorithm with Thompson sampling, a Bayesian Linear Regression reward function, and full pooling. We fit the prior for the RL algorithm using the ROBAS 2 dataset.

\subsection{Feature Space of the RL Algorithm}
\label{app:rl_alg_features}
$S_{i,t} \in \mathbb{R}^d$ represents the $i$th participant's state at decision time $t$, where $d$ is the number of features describing the participant's state.

\paragraph{\bf{Advantage Feature Space}}
$f(S_{i,t}) \in \mathbb{R}^4$ denotes the feature space used to predict the advantage (i.e. the immediate treatment effect) which is the following:
\begin{enumerate}
    \item Bias / Intercept Term $\in \mathbb{R}$
    \item Time of Day (Morning/Evening) $\in \{0, 1\}$
    \item Exponential Average of Brushing Over Past 7 Days (Normalized) $\in \mathbb{R}$
    \item Exponential Average of Messages Sent Over Past 7 Days $\in [0, 1]$
\end{enumerate}
The normalization procedure for Prior Day Brushing Quality is the same as the one described in Appendix~\ref{normalizations}. Features 3 and 4 are $\bar{B} = c_{\gamma}\sum_{j = 1}^{14} \gamma^{j-1} B_{i, t - j}$ and $\bar{A} = c_{\gamma}\sum_{j = 1}^{14} \gamma^{j-1} A_{i, t - j}$ respectively, the same $\bar{B}, \bar{A}$ used in the cost term of the reward.

\paragraph{\bf{Baseline Feature Space}}
$m(S_{i,t}) \in \mathbb{R}^5$ denotes the feature space used to predict the baseline reward function which contains all the above covariates and the following:

\begin{enumerate}[resume]
    \item Weekend Indicator (Weekday/Weekend) $\in \{0, 1\}$
\end{enumerate}

The feature space used by the RL algorithm candidates is different than the feature space used to model the reward in the simulation environments in order to test the robustness of the RL algorithm despite having a misspecified reward model.

\subsection{Bayesian Linear Regression with Action Centering}

Recall that our model for the reward is a Bayesian Linear Regression (BLR) model with action centering. Since we consider an algorithm that does full pooling (clustering with cluster size $N$), the algorithm is shared between all users in the study and therefore shares parameters $\alpha_0, \alpha_1, \beta$.

\begin{equation}
\label{eqn:blr}
    R_{i, t} = m(S_{i, t})^T \alpha_0 + \pi_{i,t} f(S_{i, t})^T \alpha_1 + (A_{i, t} - \pi_{i, t}) f(S_{i, t})^T \beta + \epsilon_{i,t}
\end{equation}
where $\pi_{i,t}$ is the probability that the RL algorithm selects action $A_{i,t} = 1$ in state $S_{i,t}$ for participant $i$ at decision time $t$. $\epsilon_{i,t} \sim \mathcal{N}(0, \sigma^2)$ and there are priors on $\alpha_{0} \sim \mathcal{N}(\mu_{\alpha_0}, \Sigma_{\alpha_0})$, $\alpha_{1} \sim \mathcal{N}(\mu_{\beta}, \Sigma_{\beta})$, $\beta \sim \mathcal{N}(\mu_{\beta}, \Sigma_{\beta})$. We discuss how we set informative prior values for $\mu_{\alpha_0}, \Sigma_{\alpha_0}, \mu_{\beta}, \Sigma_{\beta}, \sigma^2$ using the ROBAS 2 data set in Appendix~\ref{fitting_prior}.

\subsection{Fitting the Prior for the Reward Function}

\label{fitting_prior}
We use the ROBAS 2 dataset to inform priors on $\alpha_0, \alpha_1, \beta$ as well as setting the noise variance term $\sigma^2$ in Equation~\eqref{eqn:blr}. We follow the procedure as described in \cite{DBLP:journals/corr/abs-1909-03539}. Values are shown in Table~\ref{informative_prior_vals}.
\begin{table*}[ht]
    \centering
    \begin{tabular}{c|c}
        Parameter & Value \\
        \hline
        $\sigma^2$ & 3396.449 \\
        $\mu_{\alpha_0}$ & $[0, 4.925, 0, 0, 82.209]^T$ \\
        $\Sigma_{\alpha_0}$ & $\text{diag}(29.090^2, 30.186^2, 29.624^2, 12.989^2, 46.240^2)$ \\
        $\mu_{\beta}$ & $[0, 0, 0, 0]^T$ \\
        $\Sigma_{\beta}$ & $29.624^2$  $\cdot I_4$ \\
    \end{tabular}
    \caption{\textbf{Prior values for the RL algorithm informed using the ROBAS 2 data set.} Values are rounded to the nearest 3 decimal places. After performing Generalized Estimating Equations' (GEE) analysis, we found the Prior Day Total Brushing Duration feature and the bias term to be significant.}
    \label{informative_prior_vals}
\end{table*}

\subsubsection{Fitting $\sigma^2$}
$\sigma^2$ is set using ROBAS 2 and fixed for the entire study. To choose the value, we fit a separate Generalized Estimating Equations' (GEE) linear regression model per user with the feature space of the RL algorithm described in Appendix~\ref{app:rl_alg_features} and set $\sigma^2$ to be the average of the empirical variance of the residuals for each user model across the 32 user models. Notice that ROBAS 3 had a more sophisticated sensory suite than ROBAS 2, so while the scientific team was interesting in maximizing users' brushing qualities using the RL algorithm, we only have brushing durations from ROBAS 2. However, brushing duration is a reasonable guess for brushing quality, especially because average pressure duration is small (definition of brushing quality is found in Appendix~\ref{brush_quality_def}). Therefore, when performing GEE analysis, the target is brushing duration.

\subsubsection{Fitting $\mu_{\alpha_0}, \Sigma_{\alpha_0}, \mu_{\beta}, \Sigma_{\beta}$}
To set values for $\mu_{\alpha_0}, \Sigma_{\alpha_0}$, we follow the procedure described in \cite{DBLP:journals/corr/abs-1909-03539}. We do the following: 1) conduct GEE regression analyses using all participant's data in ROBAS 2 in order to assess the significance of each feature; 2a) for features that are significant in the GEE analysis, the prior mean is set to be the point estimate found from GEE analysis and the prior standard deviation is set to be the standard deviation across user models from GEE analysis; 2b) for features that are not significant, the prior mean is 0 and we shrink the standard deviation by half. For feature 3. Exponential Average of Brushing Over Past 7 Days, we imputed that value with the average of past brushing obtained so far for the first week. Recall that ROBAS 2 had no data under $A_{i, t} = 1$, so for feature 4. Exponential Average of Messages Sent Over Past 7 Days, we set the prior mean to 0 and set the standard deviation to the average prior SD of the other features. Notice that $\Sigma_{\alpha_0}$ is a diagonal matrix where the diagonal values are the prior variances described above. 

Again, because ROBAS 2 only had no data under $A_{i, t} = 1$ so we cannot use the same procedure as described above to set $\mu_{\beta}, \Sigma_{\beta}$. Instead, we set $\mu_{\beta} = \mathbf{0}$ and set $\Sigma_{\beta} = \sigma_{\beta}^2I$ (diagonal matrix where each entry on the diagonal is $\sigma_{\beta}^2$). $\sigma_{\beta}$ is equal to the average prior SD of the other features discussed above.

\subsection{Posterior Updating}
\label{posterio_update:blr}
During the update step, the reward approximating function will update the posterior with newly collected data. Since we chose a full pooling algorithm, the algorithm will update the posterior using data shared between all users in the study. 
Here are the procedures for how the Bayesian linear regression model performs posterior updating. 

Recall from the main text that we simulate users incrementally joining the study. We use $t \in [1 \colon T]$ to index the $t^{\mathrm{th}}$ decision time for a given user. We use $\tau$ to index the $\tau^{\mathrm{th}}$ update time of the algorithm (the algorithm is only updated once a week); we use $\tau(i,t)$ to denote the function that takes in the user index $i$ and the user decision time $t$ and outputs the number of full weeks since the study started.
Suppose we are selecting actions for decision time $t$ for a user $i$.
Let $\phi(S_{i, t}, A_{i, t}) = [m(S_{i, t})$, $\pi_{i, t}f(S_{i, t})$, $(A_{i, t} - \pi_{i, t})f(S_{i, t})]^\top$ be the joint feature vector and $\theta = [\alpha_0, \alpha_1, \beta]$ be the joint weight vector. Notice that Equation~\ref{eqn:blr} can be vectorized in the form: $R_{i, t} = \phi(S_{i, t}, A_{i, t})^\top \theta + \epsilon$. Notice that $\theta$ is shared amongst users because we are performing full pooling. Let $\Phi_{1:\tau(i,t)} \in \mathbb{R}^{K \cdot 5 + 4 + 4}$ be matrix of all users' data that have been collected in the study up to update-time $\tau(i,t)$, specifically, it is the matrix of all stacked vectors $\{\phi(S_{i, t}, A_{i, t}) \}$, where $K$ is the total number of user decision times in the shared history (since we simulate incremental recruitment this is not just $N \cdot (t-1)$).
Let $\mathbf{R}_{1:\tau(i,t)} \in \mathbb{R}^K$ be a vector of stacked rewards $\{ R_{i, t}\}$, a vector of all users' rewards that have been collected in the study up to update-time $\tau$.

Recall that we have normal priors on $\theta$ where $\theta \sim \mathcal{N}(\mu_{\text{prior}}, \Sigma_{\text{prior}})$, $\mu_{\text{prior}} = [\mu_{\alpha_0}, \mu_{\beta}, \mu_{\beta}] \in \mathbb{R}^{4+4+5}$ and $\Sigma_{\text{prior}} = \text{diag}(\Sigma_{\alpha_0}, \Sigma_{\beta}, \Sigma_{\beta})$. At the update time $\tau$, $p(\theta | H_{\tau})$, the posterior distribution of the weights given current history $H_{\tau}$ for all users who have entered the study, is conjugate and normal. 

$$
\theta | H_{\tau} \sim \mathcal{N}(\mu_{\tau}^{post}, \Sigma_{\tau}^{post}) 
$$
$$\Sigma_{\tau}^{post} = \bigg(\frac{1}{\sigma^2}\Phi_{1:\tau}^T \Phi_{1:\tau} + \Sigma_{prior}^{-1}\bigg)^{-1}
$$
$$\mu_{\tau}^{post} = \Sigma_{\tau}^{post} \bigg(\frac{1}{\sigma^2}\Phi_{1:\tau}^T \mathbf{R}_{1:\tau} + \Sigma_{prior}^{-1}\mu_{prior} \bigg)$$


\subsection{Action Selection}
\label{action_selection}
Our action selection scheme at decision time selects action $A_{i, t} \sim \text{Bern}(\pi_{i, t})$ where $\pi_{i,t} = \mathrm{clip} \left( \tilde{\pi}_{i,t} \right)$. $\mathrm{clip}$ is the clipping function defined in Equation~\eqref{eqn:clipping_function} and $\tilde{\pi}_{i,t}$ is the posterior probability that $A_{i, t} = 1$ is defined in Appendix~\ref{action_selection_posterior_sampling}.

\subsubsection{{Posterior Sampling}}
\label{action_selection_posterior_sampling}
Based on the Bayesian linear regression model of the reward, specified by Equation~\eqref{eqn:blr}:
$$
\tilde{\pi}_{i,t} = \text{Pr}_{\tilde{\beta} \sim \mathcal{N}(\mu^{post}_{\tau(i,t)}, \Sigma^{post}_{\tau(i,t)}) } \left\{f(S_{i, t})^T \tilde{\beta} > 0 \big| S_{i, t}, H_{\tau(i,t)} \right\}
$$
Note that the randomness in the probability above is only over the draw of $\tilde{\beta}$ from the posterior distribution.

\subsubsection{{Clipping to Form Action Selection Probabilities}}
\label{clipping_function}

Since we want to facilitate after-study analyses, we clip action selection probabilities using the action clipping function for some $\pi_{\min}, \pi_{\max}$ where $0 < \pi_{\min} \leq \pi_{\max} < 1$ is chosen by the scientific team:
\begin{equation}
\label{eqn:clipping_function}
    \mathrm{clip}(\pi) = \min(\pi_{\max}, \max(\pi, \pi_{\min})) \in [\pi_{\min}, \pi_{\max}]
\end{equation}
For our simulations, $\pi_{\min}=0.1, \pi_{\max}=0.9$.

\subsection{Relating the Cost Term to the Bellman Equation}
\label{relation_to_bellman}
Throughout we will use $\mathcal{Q}^{\pi}(s, a)$ to denote the value (immediate and future value) of taking action $a$ in state $s$, under policy $\pi$ (a mapping from $\mathcal{S}$ to distributions over the actions).
\begin{equation*}
     \mathcal{Q}^{\pi}(s,a) := \mathbb{E}_{\pi} \big[ R_t + \gamma R_{t+1} + \gamma^2 R_{t+2} + \dots \big| S_t = s, A_t = a \big]
\end{equation*}
By the Bellman equation,
\begin{equation*}
     \mathcal{Q}^{\pi}(s,a) = \mathbb{E} \big[ R_t \big| S_t = s, A_t = a \big] + \gamma \mathbb{E}_{\pi} \left[ \mathcal{Q}^{\pi} (S_{t+1}, A_{t+1}) \big| S_t = s, A_t = a \right]
\end{equation*}
\begin{equation*}
     = \mathbb{E} \big[ R_t \big| S_t = s, A_t = a \big] + \gamma \sum_{s' \in \mathcal{S}} p(S_{t+1} = s' | s, a) \sum_{a' \in \{0, 1\} } \pi(s, a') \mathcal{Q}^{\pi} (s', a')
\end{equation*}
Let $\pi_{\mathrm{MDP}}^*$ be the optimal policy in the MDP environment. It selects actions as follows: 
\begin{equation*}
  \pi_{\mathrm{MDP}}^*(s, 1) = \mathbb{I} \big\{ \mathcal{Q}^{\pi_{\mathrm{MDP}}^*}(s,a=1) - \mathcal{Q}^{\pi_{\mathrm{MDP}}^*}(s,a=0) > 0 \big\}
\end{equation*}
\begin{equation}
    \label{eqn:MDPeta}
    = \mathbb{I} \bigg\{ \mathbb{E}[R_{t}| S_t = s, A_t = 1] - \mathbb{E}[R_{t}| S_t = s, A_t = 0] > \eta(s) \bigg\}
\end{equation}
where $\eta(s)$ represents the difference in future value between taking action $0$ and $1$, i.e.,
\begin{equation*}
    \eta(s) = \gamma \mathbb{E}_{\pi_{\mathrm{MDP}}^*} \left[ \mathcal{Q}^{\pi_{\mathrm{MDP}}^*} (S_{t+1}, A_{t+1}) \big| S_t = s, A_t = 0 \right]
    - \gamma \mathbb{E}_{\pi_{\mathrm{MDP}}^*} \left[ \mathcal{Q}^{\pi_{\mathrm{MDP}}^*} (S_{t+1}, A_{t+1}) \big| S_t = s, A_t = 1 \right]
\end{equation*}
\begin{equation*}
\begin{split}
    = \gamma\sum_{s' \in \mathcal{S}} p(S_{t+1} = s' | s, 0) \sum_{a' \in \{0, 1\} } \pi_{\mathrm{MDP}}^*(s, a') \mathcal{Q}^{\pi_{\mathrm{MDP}}^*} (s', a') \\
    - \gamma\sum_{s' \in \mathcal{S}} p(S_{t+1} = s' | s, 1) \sum_{a' \in \{0, 1\} } \pi_{\mathrm{MDP}}^*(s, a') \mathcal{Q}^{\pi_{\mathrm{MDP}}^*} (s', a')
\end{split}
\end{equation*}
We can define the value of a policy $\pi$ as 
\begin{equation*}
     V^{\pi}(s) := \mathbb{E}_{\pi} \big[ R_t + \gamma R_{t+1} + \gamma^2 R_{t+2} + \dots \big| S_t = s \big].
\end{equation*}
Given this new notation, we can also simplify 
\begin{equation*}
    \eta(s) = \gamma \mathbb{E} \left[ V^{\pi_{\mathrm{MDP}}^*} (S_{t+1}) \big| S_t = s, A_t = 0 \right]
    - \gamma \mathbb{E} \left[ V^{\pi_{\mathrm{MDP}}^*} (S_{t+1}) \big| S_t = s, A_t = 1 \right]
\end{equation*}

\paragraph{Relationship to Contextual Bandits}
Notice that in a contextual bandit environment, the state transition probabilities are exogenous, so $p(S_{t+1} = s' | s, a=1) = p(S_{t+1} = s' | s, a=0)$. This means that $\eta(s) = 0$ in contextual bandit environments. The optimal contextual bandit policy $\pi_{\mathrm{CB}}^*$ selections actions \begin{equation}
    \label{eqn:optimalCB}
    \pi_{\mathrm{CB}}^*(s, 1) = \mathbb{I} \left\{ \mathbb{E} \big[ R_t \big| S_t = s, A_t = 1 \big] - \mathbb{E} \big[ R_t \big| S_t = s, A_t = 0 \big] > 0 \right\}
\end{equation}
This reflects how posterior sampling algorithms in contextual bandit environments select actions as follows:
\begin{equation}
    \label{eqn:contextTS}
    \pi_t(s, 1) = \mathbb{P}_{ \theta \sim p(\theta | \mathrm{Data}) } \bigg[ r_{\theta}(s, a=1) -r_{\theta}(s, a=0) > 0 \bigg| \mathrm{Data} \bigg]
\end{equation}
where $r_{\theta}(s,a)$ is the model we use for the mean reward $\mathbb{E} \big[ R_t \big| S_t = s, A_t = a \big]$ parameterized by $\theta$. Posterior sampling involves putting a prior on $\theta$ and posterior $p(\theta| \mathrm{Data})$ represents the posterior distribution of $\theta$. Above we can think of the posterior sampling contextual bandit algorithm as setting $\eta(s) = 0$, as in this environment there are no delayed effects of actions, i.e., actions cannot lead one to encounter unfavorable states in the future with higher probability.

\paragraph{Relationship to Surrogate Rewards}
Consider the surrogate reward we designed in the paper: $R_t = Q_t - C_t$. According to Equation \eqref{eqn:optimalCB}, the optimal contextual bandit policy selects actions according to 
\begin{equation*}
    \pi_{\mathrm{CB}}^*(s, 1) = \mathbb{I} \left\{ \mathbb{E} \big[ R_t \big| S_t = s, A_t = 1 \big] - \mathbb{E} \big[ R_t \big| S_t = s, A_t = 0 \big] > 0 \right\}
\end{equation*}
Since $R_t = Q_t - C_t$, we can equivalently say that the optimal contextual bandit policy is
\begin{equation}
    \label{eqn:CBeta}
    \pi_{\mathrm{CB}}^*(s, 1) = \mathbb{I} \bigg\{ \mathbb{E} \big[ Q_t \big| S_t = s, A_t = 1 \big] - \mathbb{E} \big[ Q_t \big| S_t = s, A_t = 0 \big] > \tilde{\eta}(s) \bigg\}
\end{equation}
where $\tilde{\eta}(s) = \mathbb{E} \big[ C_t \big| S_t = s, A_t = 1 \big] - \mathbb{E} \big[ C_t \big| S_t = s, A_t = 0 \big]$. 
The $\tilde{\eta}(s)$ term above in Equation \eqref{eqn:CBeta} is mimicking the delayed effects of actions term in the equation for the optimal MDP policy, Equation \eqref{eqn:MDPeta}. Recall in our setting $C_t$ is a function of $S_t, A_t$, where $C_t = 0$ if $A_t = 0$. This implies that $\tilde{\eta}(s) = \mathbb{E} \big[ C_t \big| S_t = s, A_t = 1 \big] = C_t$.

When we use a posterior sampling contextual bandit algorithm with a surrogate reward to select actions, we can think of the posterior sampling procedure from Equation \eqref{eqn:contextTS} as follows:
\begin{equation*}
    \pi_t(s, 1) = \mathbb{P}_{ \theta \sim p(\theta | \mathrm{Data}) } \bigg[ q_{\theta}(s, a=1) - q_{\theta}(s, a=0) > \tilde{\eta}(s) \bigg| \mathrm{Data} \bigg]
\end{equation*}
where $q_{\theta}(s,a)$ is the model we use for the mean brushing quality $\mathbb{E} \big[ Q_t \big| S_t = s, A_t = a \big]$ parameterized by $\theta$ (recall $C_t$ is a function of only $S_t$ and $A_t$, not $\theta$).
Therefore adding the cost term helps us capture the delayed effects of actions as a full MDP-based algorithm would, but in a simple way. This allows us to continue to use bandit algorithms in settings where full RL is less feasible.

\end{document}